\documentclass[10pt,twocolumn,letterpaper]{article}

\usepackage{cvpr}

\usepackage{multirow}
\usepackage{threeparttable}
\usepackage{array}

\definecolor{cvprblue}{rgb}{0.21,0.49,0.74}
\usepackage[pagebackref,breaklinks,colorlinks,allcolors=cvprblue]{hyperref}
\usepackage{tikz}
\usetikzlibrary{arrows.meta,calc}

\def\paperID{*****}
\def\confName{CVPR}
\def\confYear{2026}

\title{FROST-STA: Frozen Dense Features for the Ego4D Short-Term Object Interaction Anticipation}

\author{
Chaoyang Wang$^{1}$, Lexuan Xu$^{1}$\\
$^{1}$Beihang University\\
{\tt\small \{chaoyangwang, xulexuan\}@buaa.edu.cn}
}

\begin{document}
\maketitle
\begin{abstract}
Short-term anticipation in egocentric video requires more than recognizing the current scene: a system must infer which object the camera wearer will contact, which action will follow, and how soon the contact will happen.
This report describes \textbf{FROST-STA}, our submission to the Ego4D Short-Term Object Interaction Anticipation (STA) Challenge at EgoVis 2026.
For each query time, the model produces a ranked set of structured hypotheses containing an active-object box, noun label, verb label, time-to-contact (TTC), and confidence.
FROST-STA builds on the V-JEPA 2.1 STA evaluation protocol, but adapts it to the challenge by using object-centric decoding, multi-head prediction, and a submission-oriented training and ensembling recipe.
We keep the V-JEPA 2.1 ViT-G backbone fixed and extract two dense token streams: video tokens from a short clip resized to 384 pixels before the query, and image tokens from the last observed high-resolution frame.
A compact alignment module, consisting of an attentive probe and frame-guided temporal pooling, maps the clip representation onto the spatial reference of the final frame before fusing it with image features.
The fused maps are decoded by Faster R-CNN-style STA heads that estimate box offsets, nouns, verbs, TTC values, and interaction quality.
For the final leaderboard entry, we train for 25 epochs with the official training split plus additional permitted validation annotations, and combine predictions across eight heads and checkpoints from epochs 15-25.
FROST-STA obtains 5.13 Overall Top-5 mAP on the official test server, ranking second in the challenge and showing that frozen dense image-video features can serve as a strong basis for object-level interaction forecasting.
\end{abstract}

\section{Introduction}
\label{sec:intro}

\begin{figure*}[t]
  \centering
  \includegraphics[width=\linewidth]{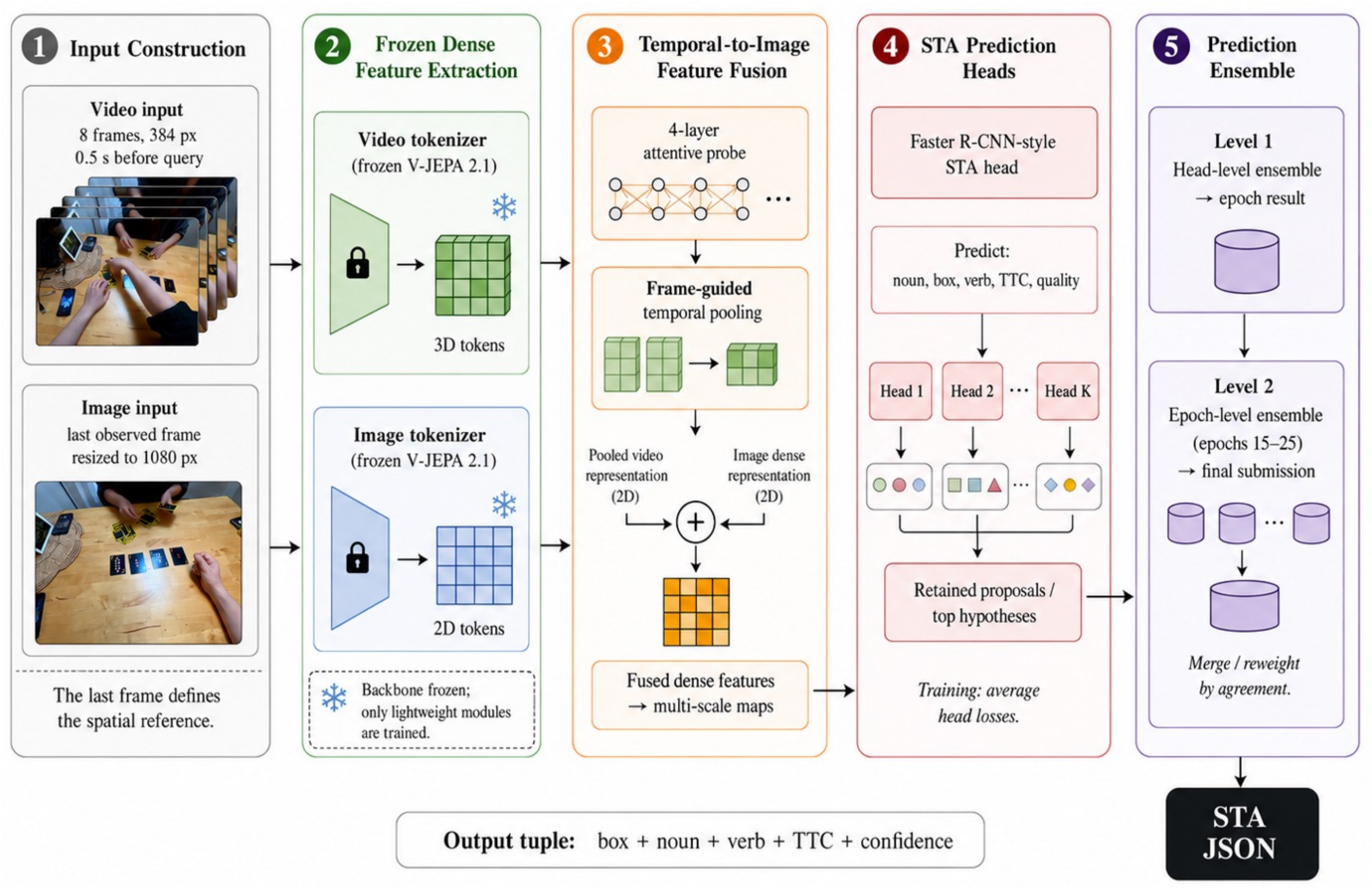}
  \caption{FROST-STA pipeline. A short video prefix and the last observed frame are processed by frozen V-JEPA 2.1 tokenizers. The video stream is compressed and aligned to the final-frame coordinate system by attentive probing and frame-guided temporal pooling. The fused dense representation is decoded by eight STA heads, and a two-stage ensemble over heads and checkpoints produces the submitted JSON predictions.}
  \label{fig:method}
\end{figure*}

Egocentric video is a natural sensing modality for agents that must act before a human-object interaction actually starts.
The Ego4D STA task formalizes this requirement as an object-level forecasting problem~\cite{grauman2022ego4d}.
At a query timestamp $t$ in an untrimmed egocentric video, only the visual evidence before $t$ is available.
A submitted model must return multiple ranked hypotheses for the next interaction, where each hypothesis specifies the box of the object that will become active, its noun class, the future verb, the TTC, and a confidence value.
The ``active'' object is defined by the forthcoming contact with the camera wearer, either directly with the hand or indirectly through a held tool.
Accurate predictions of this form are relevant for proactive assistance, warning systems, and robotic collaboration, because the system can prepare before physical contact occurs.

The task is demanding because the evaluation couples several sources of error into one structured output.
A candidate object may occupy only a few pixels, may be partly outside the field of view, or may be visually confusable with nearby items.
The model also has to decide which object is intended for the next action, not simply which objects are visible.
Finally, the TTC estimate is sensitive to brief motion and hand-object context in the last moments before the query.
Since the main leaderboard metric, Overall Top-5 mAP, accepts a detection only when localization, noun, verb, and timing constraints are jointly satisfied, progress on one component alone does not necessarily translate into a better final score.
This motivates a design that keeps high-resolution spatial evidence while still incorporating the recent temporal context.

Previous STA systems provide useful guidance for this design choice.
Faster R-CNN-style proposal mechanisms offer reliable object localization priors~\cite{ren2015faster}; SlowFast video networks provide a way to encode short-term dynamics~\cite{feichtenhofer2019slowfast}; and StillFast demonstrates that the last frame and the preceding clip can be combined effectively for short-horizon interaction anticipation~\cite{ragusa2023stillfast}.
More recently, V-JEPA 2.1 reported strong dense representations for STA when a frozen video-image backbone is paired with lightweight probing and detection heads~\cite{murlabadia2026vjepa21}.
Related frame-guided temporal pooling mechanisms, such as those used in STAformer-style models, further suggest that aligning clip features to the spatial layout of the final observed frame is beneficial~\cite{murlabadia2024affttention}.

Beyond these architectural cues, recent egocentric anticipation works by Chu et al. have shown impressive progress across both long-term and short-term settings. Their Ego4D LTA report and intention-guided INSIGHT framework highlight the value of intent-aware reasoning, while the JFAA and VISTA reports further demonstrate the strength of frozen future-predictive representations, lightweight probing/fusion, and robust ensembling in top-performing challenge systems \cite{chu2025ego4dlta,chu2026insight,chu2026jfaa,chu2026vista}.

We therefore develop \textbf{FROST-STA}, a Frozen Representation and Object-centric Short-Term Anticipation system for the Ego4D STA Challenge 2026.
As illustrated in Figure~\ref{fig:method}, FROST-STA uses V-JEPA 2.1 as a frozen dense feature extractor rather than as an end-to-end fine-tuned video backbone.
The model processes two inputs with different roles: a short, low-resolution clip carries temporal and hand-motion cues, whereas the last high-resolution frame provides the coordinate system and object details needed for box prediction.
A lightweight temporal-to-image fusion block aligns the video tokens with the image token grid, after which Faster R-CNN-style STA heads decode object-level predictions.

The final system also emphasizes robustness at inference time.
Within a checkpoint, eight parallel STA heads generate complementary hypotheses from the same fused representation.
Across training, we additionally merge epoch-level outputs from epochs 15-25.
Using the official training data together with permitted validation annotations, this recipe achieves 5.13 Overall Top-5 mAP on the official test leaderboard and ranks second in the Ego4D STA Challenge at EgoVis 2026.

\section{Method}
\label{sec:method}

Figure~\ref{fig:method} gives an overview of FROST-STA.
The system can be viewed as five stages: building query-specific inputs, extracting frozen dense features, aligning the temporal stream to the final frame, decoding STA hypotheses, and aggregating predictions.
Only the lightweight probe, fusion layers, and prediction heads are optimized for the challenge; the V-JEPA 2.1 backbone remains frozen throughout training.

\subsection{Query Inputs}

For every query timestamp, we construct an image-video pair following the input convention of the V-JEPA 2.1 STA protocol~\cite{murlabadia2026vjepa21}.
The temporal input contains 8 frames sampled from the 0.5-second interval immediately preceding the query and resized to 384 pixels.
The spatial input is the final observed frame, resized to 1080 pixels.
This frame defines the coordinate system in which the predicted active-object boxes are reported, while the short clip supplies the motion and interaction context that is not visible from a single image alone.

\subsection{Frozen V-JEPA 2.1 Token Streams}

FROST-STA uses the V-JEPA 2.1 ViT-G model as a dense, non-trainable representation backbone~\cite{murlabadia2026vjepa21}.
The high-resolution last frame is encoded by the image tokenizer, and the sampled video clip is encoded by the video tokenizer.
Because V-JEPA 2.1 uses modality-specific patchification and embeddings, the two streams preserve different information: the image stream emphasizes fine spatial appearance, whereas the video stream captures local temporal evolution.
We take the final-layer dense tokens from both streams and pass them to the task-specific modules without updating the backbone parameters.

Keeping the representation extractor fixed has two practical advantages for the challenge setting.
First, it reduces the amount of trainable computation and makes training stable when the available STA supervision is limited.
Second, it lets the downstream model exploit the dense spatial and temporal structure learned by the large self-supervised encoder while concentrating the supervised learning capacity on alignment and object-level prediction.

\subsection{Clip-to-Frame Alignment and Fusion}

The video tokens are three-dimensional and are not immediately expressed in the coordinate system of the final observed frame.
To bridge this gap, we place a four-layer attentive probe on top of the frozen video tokens, following the V-JEPA 2.1 STA evaluation setup~\cite{murlabadia2026vjepa21}.
The probed tokens are then reduced by frame-guided temporal pooling, which converts the temporal token volume into a two-dimensional representation aligned with the last-frame grid.
This step is conceptually related to the frame-guided pooling strategy used in STAformer-style STA models~\cite{murlabadia2024affttention}.

After alignment, the pooled video map is added to the dense image map.
The fused map is converted into four feature scales before being fed to the detection head, following the common feature-pyramid principle used in object detection~\cite{lin2017feature}.
In this way, the decoder receives both precise appearance information from the last frame and short-term hand-object evidence from the preceding clip.

\subsection{Object-Level STA Decoder}

The multi-scale fused features are decoded with a Faster R-CNN-style STA head~\cite{ren2015faster}, in line with the detection components used by StillFast and the V-JEPA 2.1 STA evaluation pipeline~\cite{ragusa2023stillfast,murlabadia2026vjepa21}.
For each retained proposal, the decoder predicts class-specific box refinements, noun logits over the Ego4D STA object vocabulary, verb logits for the future interaction, a positive TTC value, and an interaction-quality score.
The score used to rank final hypotheses combines proposal objectness, interaction quality, noun likelihood, and verb likelihood.s

We use $K$ prediction heads in parallel.
All heads consume the same fused ROI features, but each head has its own output layers and therefore produces a slightly different set of hypotheses.
During training, the losses from the heads are averaged.
At inference, each head expands retained proposals into top noun-verb combinations; the resulting hypotheses are ranked, merged, processed with class-aware non-maximum suppression, and written as the epoch-level prediction file.

\subsection{Head and Checkpoint Ensemble}

The submitted result is produced by a two-stage aggregation procedure.
The first stage merges the outputs of the eight heads within a single checkpoint.
The second stage combines the epoch-level files generated from checkpoints between epochs 15 and 25.

Aggregation is performed over complete STA hypotheses rather than boxes alone.
Two predictions are considered compatible when they agree on noun and verb labels and are close in both spatial overlap and TTC.
For each compatible group, the confidence is recomputed from the original scores and the amount of agreement across heads or checkpoints.
This procedure is inspired by confidence-weighted box fusion for object detection~\cite{solovyev2021weighted}, but extends the matching rule from a spatial box criterion to the full STA output tuple.
The ensemble reduces the variance of individual heads and checkpoints while keeping the final predictions consistent with the official spatial, semantic, and temporal matching requirements.

\section{Experiments}
\label{sec:experiments}

\begin{table*}[t]
  \centering
  \caption{Official Ego4D STA test leaderboard excerpt. The challenge ranks submissions by Overall Top-5 mAP, where larger values are better. Our entry is shown in bold.}
  \label{tab:leaderboard}
  \begin{threeparttable}
  \begin{tabular*}{\textwidth}{@{\extracolsep{\fill}}c l c c c c@{}}
    \toprule
    \# & Participant & Overall & Noun & Noun+Verb & Noun+TTC \\
    \midrule
    1 & corrine & 5.40 & 27.26 & 16.15 & 8.95 \\
    \textbf{2} & \textbf{sun0710} & \textbf{5.13} & \textbf{23.83} & \textbf{14.52} & \textbf{8.07} \\
    3 & StillFast Baseline V2 & 5.12 & 25.06 & 13.29 & 9.14 \\
    4 & Faster R-CNN + SlowFast Baseline V2 & 3.61 & 26.15 & 9.45 & 8.69 \\
    \bottomrule
  \end{tabular*}
  \end{threeparttable}
\end{table*}

\subsection{Training and Submission Setting}

For the challenge entry, FROST-STA is trained for 25 epochs.
The supervised data consist of the official Ego4D STA training split together with additional validation annotations permitted by the challenge rules.
Because those validation annotations are used for model fitting, we do not present a separate validation benchmark in this report.
All reported numbers are test-set results returned by the official challenge evaluation server.

\subsection{Metrics}

The official evaluation computes Top-5 mAP on the STA test split.
A predicted hypothesis is eligible for matching only if its active-object box overlaps the ground-truth box with IoU greater than 0.5; additional constraints depend on the metric variant.
Noun mAP checks the object label, Noun+Verb mAP checks both object and action labels, and Noun+TTC mAP checks the object label together with an absolute TTC error below 0.25 seconds.
The Overall score is stricter: the box, noun, verb, and TTC conditions must all be satisfied for a match.
This Overall Top-5 mAP is the primary ranking criterion because it measures whether a method can produce a coherent spatial, semantic, and temporal anticipation result.

\subsection{Official Test Results}

Table~\ref{tab:leaderboard} summarizes the leaderboard entries relevant to our comparison.
Our submission, \textit{sun0710}, reaches 5.13 Overall Top-5 mAP and places second.
Relative to the Faster R-CNN + SlowFast Baseline V2, the Overall score increases from 3.61 to 5.13, and Noun+Verb mAP rises from 9.45 to 14.52.
FROST-STA is also close to the StillFast Baseline V2 on the primary metric and obtains a higher Noun+Verb score.
At the same time, the Noun and Noun+TTC numbers are not the strongest among the listed entries, which indicates that category discrimination and fine-grained contact-time estimation remain the main weaknesses of our frozen-feature recipe.
Overall, the test result supports the effectiveness of combining frozen V-JEPA 2.1 dense tokens, frame-aligned temporal pooling, multi-head object-centric decoding, and checkpoint-level aggregation.

\subsection{Qualitative Diagnosis}

\begin{figure}[t]
  \centering
  \includegraphics[width=\columnwidth]{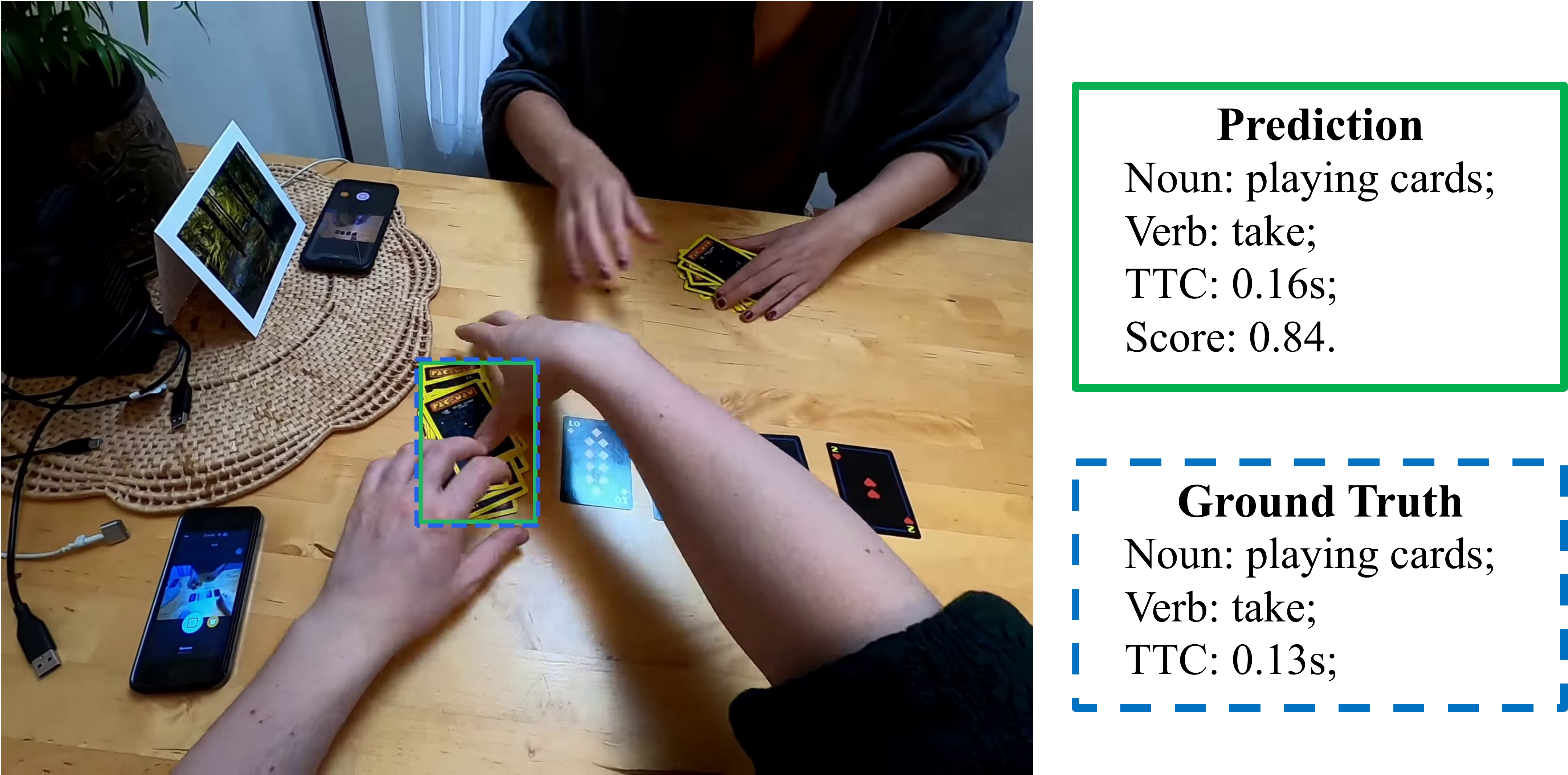}
  \caption{Qualitative success case. The predicted active-object box aligns with the annotation, and the noun, verb, and TTC prediction match the future interaction.}
  \label{fig:success_case}
\end{figure}

\begin{figure}[t]
  \centering
  \includegraphics[width=\columnwidth]{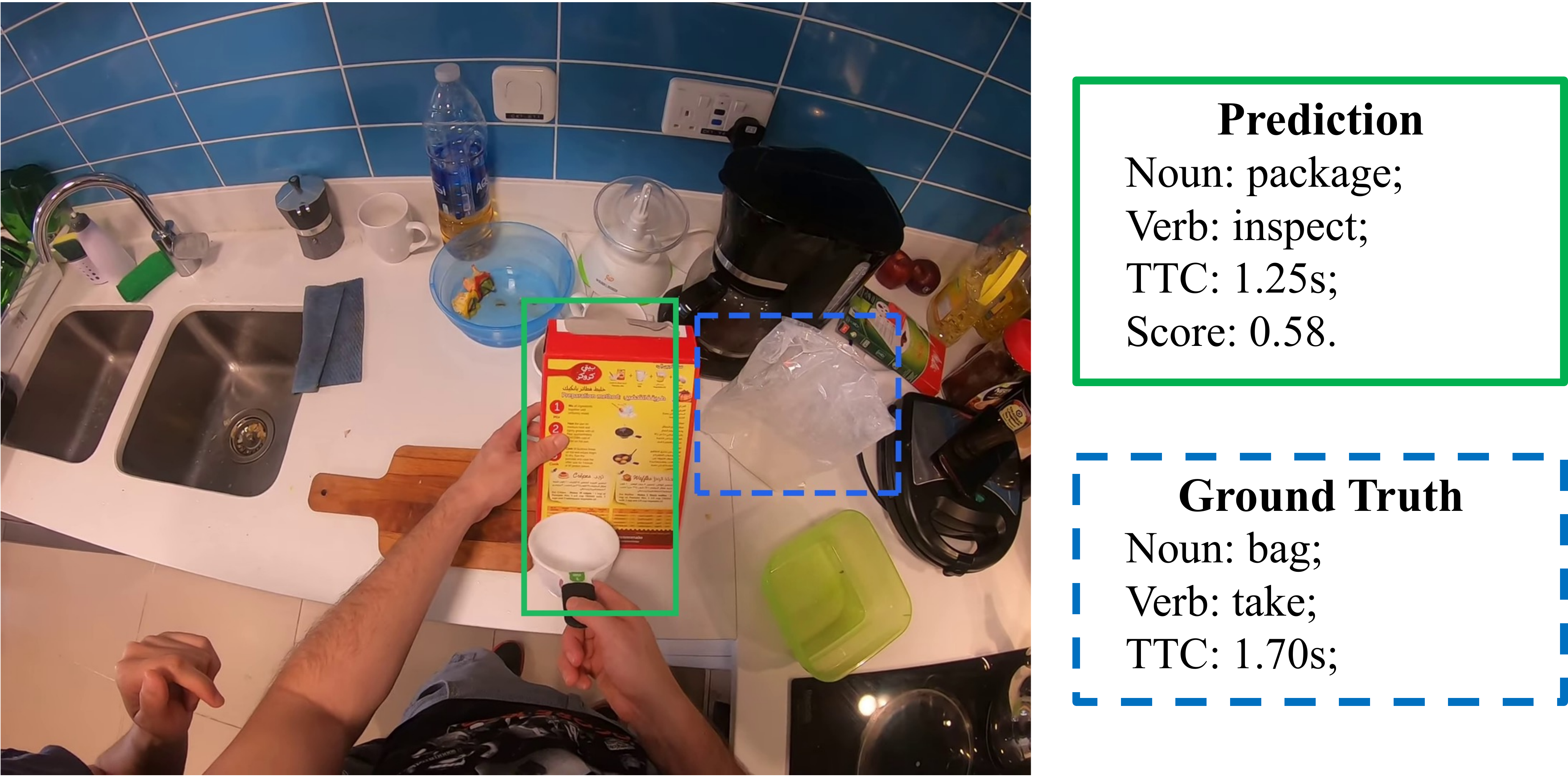}
  \caption{Qualitative failure case. The model assigns the interaction to a plausible neighboring object, while the annotated target is another object that is contacted next.}
  \label{fig:failure_case}
\end{figure}

Since the final model uses the permitted validation annotations during training, the examples in Figures~\ref{fig:success_case} and~\ref{fig:failure_case} are used only for qualitative inspection.
Each visualization shows the prediction frame, the predicted and annotated active-object boxes, and the associated noun, verb, TTC, and confidence values.
Solid green boxes correspond to model predictions, whereas dashed blue boxes mark the ground truth.

Figure~\ref{fig:success_case} presents a correct case.
The predicted box has high overlap with the annotated active object, and the predicted noun, verb, and TTC are consistent with the labeled future interaction.
This example illustrates how aligned temporal evidence can help the object decoder select the item that will soon be manipulated.

Figure~\ref{fig:failure_case} presents a typical error.
Several plausible objects appear close to the hand path, and the model assigns high confidence to a nearby distractor rather than the true future active object.
Such failures suggest that the remaining difficulty is not only detection, but also intent disambiguation under occlusion, small object size, and cluttered egocentric scenes.

\section{Conclusion}
\label{sec:conclusion}

This report introduced FROST-STA, our runner-up solution for the Ego4D STA Challenge at EgoVis 2026.
The system keeps V-JEPA 2.1 frozen, extracts dense image and video tokens, aligns the temporal stream to the last observed frame, and decodes fused multi-scale features with eight object-centric STA heads.
A lightweight ensemble over heads and checkpoints produces the final submission.
On the official test split, FROST-STA obtains 5.13 Overall Top-5 mAP and ranks second on the leaderboard.
The result indicates that frozen dense visual representations are a competitive foundation for short-term object interaction anticipation, while the remaining gaps in Noun and Noun+TTC performance point to better object-category modeling and more accurate contact-time estimation as useful future directions.

{
    \small
    \bibliographystyle{unsrt}
    \bibliography{main}
}

\end{document}